\documentclass[runningheads]{llncs}
\usepackage[T1]{fontenc}
\usepackage{graphicx}
\usepackage{csquotes}
\usepackage{amsmath,mathtools,nccmath}
\usepackage[ruled,linesnumbered]{algorithm2e}
\usepackage{algorithmic}
\usepackage{setspace}
\usepackage[inline]{enumitem}
\SetAlFnt{\footnotesize}
\usepackage{svg}
\usepackage[off]{svg-extract}
\svgsetup{clean=true}
\usepackage[colorlinks = true,
            linkcolor = blue,
            urlcolor  = blue,
            citecolor = blue,
            anchorcolor = blue]{hyperref}
\hypersetup{breaklinks=true}
\usepackage[misc]{ifsym}
\begin{document}
\title{HKD-SHO: A hybrid smart home system based on knowledge-based and data-driven services}
%
%
\author{Mingming Qiu\inst{1,2}\textsuperscript{(\Letter)} \and
Elie Najm\inst{1} \and
Rémi Sharrock\inst{1}\and
Bruno Traverson\inst{2}}

\authorrunning{M. Qiu et al.}
%

%
\institute{Télécom Paris, Palaiseau, France\\
\email{\{Mingming.Qiu, Elie.Najm, Remi.Sharrock\}@telecom-paris.fr}\\\and
EDF R\&D, Palaiseau, France\\
\email{\{Bruno.Traverson\}@edf.fr}}

\maketitle              
\begin{abstract}
A smart home is realized by setting up various services. Several methods have been proposed to create smart home services, which can be divided into knowledge-based and data-driven approaches. However, knowledge-based approaches usually require manual input from the inhabitant, which can be complicated if the physical phenomena of the concerned environment states are complex, and the inhabitant does not know how to adjust related actuators to achieve the target values of the states monitored by services. Moreover, machine learning-based data-driven approaches that we are interested in are like black boxes and cannot show the inhabitant in which situations certain services proposed certain actuators' states. To solve these problems, we propose a hybrid system called HKD-SHO (Hybrid Knowledge-based and Data-driven services based Smart HOme system), where knowledge-based and machine learning-based data-driven services are profitably integrated. The principal advantage is that it inherits the explicability of knowledge-based services and the dynamism of data-driven services. We compare HKD-SHO with several systems for creating dynamic smart home services, and the results show the better performance of HKD-SHO.

\end{abstract}
\section{Introduction}

The intelligence of a smart home is realized by creating various services. Depending on contexts and problems, different concepts of services are defined. For example, \cite{chaki2020fine} considers each device as a service. \cite{sun2014conflict} implicitly expresses a service as an action performed when certain conditions are met. \cite{ma2016detection} describes a service as a target of functions triggered under certain conditions. In our study, we define that each smart home service manages a particular state, called a monitored state, by instructing actuators to perform appropriate actions after considering the environment states sensed by sensors. Existing methods to create smart home services belong to either knowledge-based or data-driven approaches.

Knowledge-based approaches create smart home services by setting a set of rules. Therefore, knowledge-based services can show the inhabitant in which situations they have proposed certain actuators' states. In addition, the inhabitant can manually create services if he knows his preferred values for monitored states and how to set actuators to achieve these values. The manual service creation process takes little time when the services are simple. However, it becomes problematic when the number of actuators is large. In addition, the services created are often static and cannot adapt to the changing environment states and inhabitant's preferences. Finally, ensuring that the created knowledge-based services can make propositions under any environment state is challenging.

Data-driven approaches make strategic decisions based on data analysis and interpretation. Machine learning methods that learn from data and make predictions based on it are at the forefront of data-driven decision-making \cite{brunton2022data}. They seek to automatically discover the patterns of systems by analyzing the provided dataset. When developing a user-friendly smart home, it is essential to consider the reactions of the inhabitant \cite{chan2008review}. The basic idea of Reinforcement learning (RL) \cite{sutton2011reinforcement}, one of the well-known machine learning algorithms, is that an artificial agent can learn the behavior patterns of the system under study by interacting with the environment. Therefore, RL-based data-driven services can consider the inhabitant's reactions to the proposed actuators' states in a smart home to adapt to the target values of monitored states.
Moreover, systems in the real world, such as smart homes, are usually high-dimensional. Thus, Deep Reinforcement Learning (DRL) \cite{franccois2018introduction} can simplify smart home system modeling. However, since DRL is a black box constantly evolving, the inhabitant cannot tell in which situations certain actuators' states have been proposed.

To overcome the shortcomings of the two approaches and maintain their advantages, we propose to combine knowledge-based and data-driven services to obtain a hybrid system called HKD-SHO (Hybrid Knowledge-based and Data-driven services based Smart HOme system). HKD-SHO has the explicability of knowledge-based services by showing users in which situations certain actuators' states have been proposed, and the dynamism of data-driven services to dynamically adapt to the changing environment states and inhabitant's preferences. Moreover, knowledge-based services can be automatically enriched through machine learning methods to avoid manual creation by the inhabitant.

In the rest of the paper, Section \ref{sec:related_work} presents existing work on creating smart home services using hybrid approaches. Sections \ref{sec:proposed_HKD-SHO system} to \ref{sec:decision_maker} introduce the structure and each component of HKD-SHO. Section \ref{sec:working_process_of_hkd-sho} shows in detail the entire working process of HKD-SHO. 
Section \ref{sec:simulated_experiment} conducts several simulated experiments to evaluate HKD-SHO. Section \ref{sec:conclusion} concludes the paper and provides some perspectives.

\section{Related work}
\label{sec:related_work}

Some work about hybrid systems has been studied within the Internet of Things or control domains. For example, \cite{wozniak2019intelligent} proposes an approach based on a neural network and predefined rules for a smart home system, using predefined rules to control the system while collecting data and then training the neural network. Once the neural network can generate new knowledge that guarantees higher efficiency, the system switches to using the neural network. 
\cite{handelman1990integrating} has a similar principle. The difference is that once the neural network makes wrong decisions, the knowledge-based system retakes the responsibility of decision-making while teaching the neural network. 
In both studies, there is only communication in the direction from the knowledge-based method to the neural network, but not vice versa.
\cite{feng2021hybrid} deigns a hybrid intelligent control for the combustion process of the CSP (Compact Strip Production) heating furnace. A fuzzy controller improves the control accuracy of furnace temperature under stable working conditions. An expert controller adjusts the temperature rapidly under fluctuating working conditions. The caloric value compensation reduces the influence of the caloric value of gas fluctuation on temperature, and an intelligent control system improves the control precision of the furnace temperature and reduces energy consumption. 
\cite{yongjian2007design} develops a hybrid intelligent control strategy integrating adaptive fuzzy control, rule-based expert control technology, and man-machine coordinating control for the rotary kiln process. Specifically, if the status of the kiln is normal, the results of the adaptive fuzzy control are used. If the status of the kiln is abnormal, the results of expert control are used. Whenever the rotary kiln status is abnormal, the results of man-machine coordinating control are applied.
Nevertheless, the above two applications do not notice that the expert system's rules are static and that the data-driven intelligent control is not explicable.

In our proposed system HKD-SHO, knowledge-based and data-driven services are profitably integrated: data-driven services can automatically enrich knowledge-based services through rule extraction, and knowledge-based services can control the learning direction of data-driven services and explain them.

\section{Proposed HKD-SHO system}
\label{sec:proposed_HKD-SHO system}

\begin{figure}[t!]
\centering
 \begin{minipage}[t]{0.55\textwidth}
     \centering
     \includegraphics[width=\textwidth]{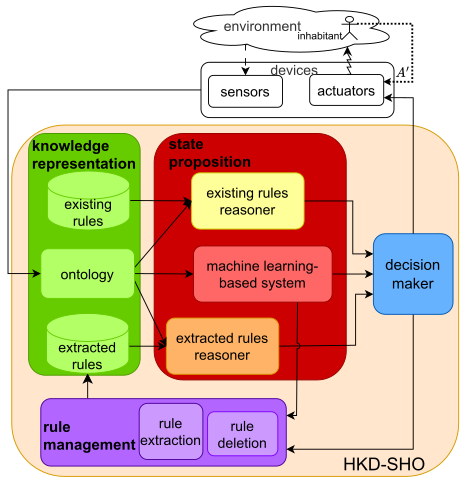}
     \caption{HKD-SHO architecture}
    \label{fig:hkd_sho}
   \end{minipage}\hspace{2mm}
   \begin{minipage}[t]{0.41\textwidth}
     \centering
     \includegraphics[width=\textwidth]{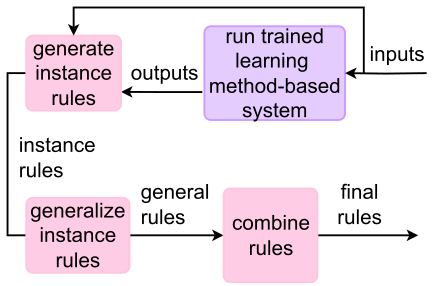}
     \caption{principle of PBRE variant}
  \label{fig:pbre}
   \end{minipage}
\end{figure}

Fig.\ref{fig:hkd_sho} shows the structure of HKD-SHO consisting of four modules: "knowledge representation", "rule management", "state proposition" and "decision maker".
The knowledge representation module contains an "ontology" and two databases that respectively stores existing and extracted rules. The ontology we use is the Smart Appliances REFerence (SAREF) ontology \cite{SAREF}, which enables semantic interoperability for smart appliances. Existing rules are written in Semantic Web Rule Language (SWRL) \cite{horrocks2004swrl}. Extracted rules are acquired by our proposed rule extraction algorithm whose working process is explained in Section \ref{sec:rule_management}.

\section{Rule management}
\label{sec:rule_management}

The "rule management" module is responsible of updating the extracted rules, including extracting and deleting these rules. The two operations are respectively completed by the "rule extraction" and "rule deletion" modules.

\subsection{Rule extraction}

To extract rules, we propose a new rule extractor, the PBRE variant, building upon PBRE \cite{qiu2022pbre}. Unlike PBRE, our variant omits the manual "refine rules" process, which removes less critical states based on evaluating the accuracy of the extracted rules without these states in the conditions, by considering all states used as inputs of the machine learning-based system as essential. Rules are dynamically extracted at each time step if the inhabitant is satisfied, rewards being positive, with propositions of the machine learning-based system. 


The PBRE variant's operational flow, depicted in Fig.\ref{fig:pbre}, generates an instance rule at each time step. This rule, capturing a specific situation, is formed by combining the states considered by the machine learning-based system and the proposed outputs using "if-then" rules, as illustrated in Eq.\ref{eq:rule_instance_1}.
\begin{flalign}
{
    \begin{aligned}\label{eq:rule_instance_1}
     &\textit{if state }x_0 \textit{ has value }x_{i_0,t},\cdots, \textit{ and }x_n \textit{ has value }x_{i_n,t}, \textit{ then actuator}\\
    &
    {a}_0 \textit{ will have state }s_{a_{0},t}, 
    \cdots, \textit{and }
    {a}_m \textit{ will have state }s_{a_{m},t}.
    \end{aligned}
      }
\end{flalign}
Next, the instance rule undergoes a generalization process by merging with existing rules that share similar conclusions and have comparable values in the conditions' states. During generalization, each state's average, minimum, and maximum values are computed based on related merging (instance) rules. Additionally, the number of occurrences, indicating how frequently an actuator's state is proposed by the machine learning-based system, is tallied. After generalization, the instance rule transforms into rules describing broader situations. Eventually, all available rules are combined, merging those with matching conclusions and overlapping range values in the conditions' states. The final extracted rule is expressed in Eq.\ref{eq:example_final_extracted_rules}.
\begin{flalign}
{
  \begin{aligned}
  \label{eq:example_final_extracted_rules}
&\textit{if state }x_0 \textit{ is between }x_{i_0,min} \textit{ and }x_{i_0,max} \textit{ and has average } x_{i_0,m},
\cdots, \textit{then}\\
&  
\textit{actuator }
{a}_0 
\textit{ will have state }s_{{a}_0} \textit{ with } \textit{frequency of occurrence }{count}_{{a}_0}, 
\cdots.
  \end{aligned}
  }
\end{flalign}

\subsection{Rule deletion}

The "rule deletion" module removes the rule extracted for proposing actuator states at time step $t-1$ from the "extracted rules" database when the inhabitant is dissatisfied with the propositions at $t$. Dissatisfaction is defined by the machine learning-based system yielding non-positive rewards for any service at $t$.

\section{State proposition}


Three mechanisms are available in the "state proposition" module to create services by proposing actuators' states after considering related environment states: existing rules reasoner, machine learning-based system, and extracted rules reasoner. Pellet \cite{sirin2007pellet}, the first sound and complete OWL-DL reasoner with extensive support for reasoning with individuals, user-defined datatypes, and debugging support for ontologies \cite{SIRIN200751}, is the existing rules reasoner applied in our study to make inferences on existing rules. Subsequent sections will introduce the remaining two mechanisms based on recently proposed techniques.

\subsection{Machine learning-based system}

Our study employs the selected SHOMA (Smart HOme-based Multi-services Architectures) machine learning-based system, specifically the RSAbA (Remove Shared Actuators-based Architecture) and EPbA (Equal Priority-based Architecture) \cite{qiu2022reinforcement}. SHOMA, a set of Multi-Agent Reinforcement Learning (MARL) architectures, is designed to create multiple smart home services without conflicts. Briefly, RSAbA and EPbA principles can be outlined using three services, denoted as $z_1, z_2,$ and $z_3$, each controlling distinct sets of actuators: $\{d_1,d_2,d_3\}$ for $z_1$, $\{d_2,d_4,d_5\}$ for $z_2$ and $\{d_2,d_3,d_6\}$ for $z_3$, where $d_2$ is shared by all three services, and $d_3$ is shared by $z_1$ and $z_3$. Also, the input states considered by each service are ${O^{z_1}}=\{{o}^{z_1}_{1},{o}^{z_1}_{2},\cdots\}$, $O^{z_2}=\{{o}^{z_2}_{1},{o}^{z_2}_{2},\cdots\}$, and $O^{z_3}=\{{o}^{z_3}_{1},{o}^{z_3}_{2},\cdots\}$.


\begin{figure}[t!]
\centering
 \begin{minipage}[t]{0.51\textwidth}
     \centering
     \includegraphics[width=\textwidth]{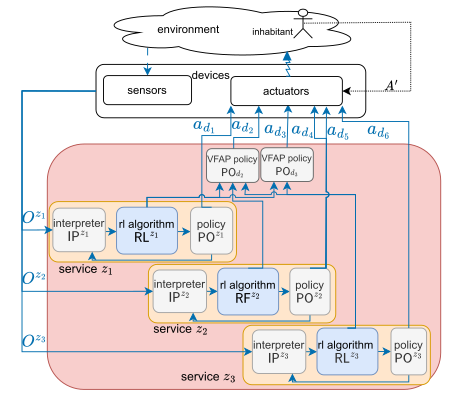}
     \caption{Architecture of EPbA}
    \label{fig:epba}
   \end{minipage}
   \begin{minipage}[t]{0.48\textwidth}
     \centering
     \includegraphics[width=\textwidth]{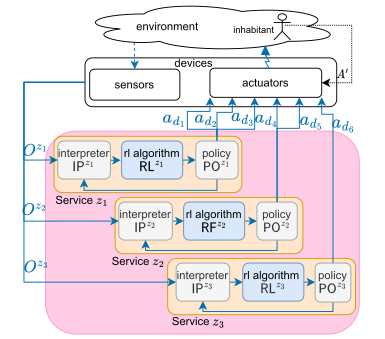}
     \caption{Architecture of RSAbA}
  \label{fig:rsaba}
   \end{minipage}
\end{figure}

\subsubsection{EPbA}

In the EPbA principle (Fig.\ref{fig:epba}), each service acts as an agent with an "interpreter" $IP$, an "RL algorithm" $RL$, and a "policy" $PO$. $IP$ is designed to extract essential states for making propositions, but in our assumption that all states are essential, $IP$ remains inactive. $RL$ proposes action quality values $Q$ for potential states of each actuator, representing the expected long-term reward for selecting a state as the new state of the associated actuator at the current time step. We employ deep Q-learning \cite{hester2018deep} for $RL$. Finally, $PO$ selects final states for each actuator using the $\epsilon$-greedy policy \cite{tokic2011value}, choosing the state with the maximum action quality value if a randomly generated value exceeds a predefined small $\epsilon$, otherwise randomly selecting states for each actuator.


The values of non-shared actuators ($d_1, d_4, d_5,$ and $d_6$) are determined by the respective services. For shared actuators ($d_2$ and $d_3$), we employ the Value Function Addition Principle (VFAP) policy \cite{sunehag2017value}, where the $Q$ values calculated by the sharing services are summed, and the state with the maximum value is chosen as the final state for that shared actuator. For instance, $Q_{d_2}$ is the sum of the action quality values proposed by $z_1, z_2$, and $z_3$ for $d_2$:
\begin{equation}
\label{eq:PbSHOMMA_action_quality_act_1}
\small{
\begin{multlined}
Q_{{d}_2}=Q^{z_1}_{{d}_2}+Q^{z_2}_{{d}_2}+Q^{z_3}_{{d}_2}.
\end{multlined}
}
\end{equation}
Thus, the selected state for $d_2$ is $a_{d_2}=argmax(Q_{{d}_2})=argmax(q^1_{{d}_2},q^2_{{d}_2},\cdots)$ 
where $argmax$ selects the index of $Q_{{d}_2}$ with the maximum value. If the inhabitant is dissatisfied with the proposed actuator states or the updated indoor monitored states, they can modify actuator states or set the target value for the monitored state by adjusting the variable $A^\prime$.


\subsubsection{RSAbA}

In contrast to EPbA, RSAbA eliminates shared actuators by allowing only one sharing service to control them. The chosen service is the one with the simplest dynamic characteristics regarding the evolution of the associated monitored state. For instance, if the evolution behavior of indoor light intensity is simpler than that of indoor temperature, the service controlling light intensity is selected. Assuming a complexity order among $z_1, z_2,$ and $z_3$ as $complexity(z_1)<complexity(z_2)<complexity(z_3)$, shared actuators $d_2$ and $d_3$ are controlled by $z_1$ as shown in Fig.\ref{fig:rsaba}.


\subsection{Extracted rules reasoner}

The "extracted rules reasoner" makes inferences on the extracted rules acquired by the PBRE variant. Its principle is shown in Algo.\ref{algo:extracted_rules_reasoner}: At some time step, we suppose that the input states are $O$, and the extracted rules are $GR$. To make an inference, we determine whether, among $GR$, there are ${GR}_2$ whose range values of the states in the conditions contain the values of $O$ (line \ref{algo:inference_variant_pbre_1_1}). If yes and the number of rules in ${GR}_2$ is 1, then the conclusion of ${GR}_2$ is the final proposition of the reasoner (lines \ref{algo:inference_variant_pbre_1_3}$\sim$\ref{algo:inference_variant_pbre_1_4}). Otherwise, we concatenate the state values of $O$ and the average values of the states in the conditions of each extracted rule in $GR$ (line \ref{algo:inference_variant_pbre_2}). Then we calculate the correlation between $O$ and each rule in $GR$ using Pearson product-moment correlation coefficients (PPMCC) \cite{benesty2009pearson} (line \ref{algo:inference_variant_pbre_3}). If there exists one rule that has the closest correlation with $O$, we select the conclusion of that rule (lines \ref{algo:inference_variant_pbre_4}$\sim$\ref{algo:inference_variant_pbre_5}). Otherwise, we select the conclusion of the rule whose conclusion has the maximum number of occurrences (line \ref{algo:inference_variant_pbre_6}). Finally, we return the selected conclusion as the proposition of the reasoner (line \ref{algo:inference_variant_pbre_7}).  

\begin{algorithm*}[t!]
\algsetup{linenosize=\tiny}
\setstretch{0.93}
\DontPrintSemicolon 
\SetAlgoLined
\KwResult{Inference result}

$GR$: available extracted rules\;
$O$: current sample containing state sets $O$\;

\SetKwFunction{FMain}{{\textbf{Inference}}}
  \SetKwProg{Fn}{Function}{:}{}
  \Fn{\FMain{\text{\upshape $GR$, $O$}}}{

  
  ${GR}_{2} \leftarrow$ select the extracted rules from GR whose ranges of states in the conditions contain the values of the states of $O$\label{algo:inference_variant_pbre_1_1}\;
  
  \lIf{$size({GR}_2)\ equals\ 1$\label{algo:inference_variant_pbre_1_3}}{
  
    select $conclusions$ of ${GR}_2$ \label{algo:inference_variant_pbre_1_4}
  }
  \Else{\label{algo:inference_variant_pbre_1_5}
    $arr$=concatenate $O$ and ${{GR}_2}.states.avg$ \label{algo:inference_variant_pbre_2}\;
    
    $corr=correlation(arr)$ \label{algo:inference_variant_pbre_3}\;

  \lIf{$\forall i,j\in corr, (i-j)<\epsilon$ \label{algo:inference_variant_pbre_4}}{
  
  select $conclusion$ of the rule which has the maximum sum of number of occurrence of the conclusions\label{algo:inference_variant_pbre_5}
  }
  \Else{
    select $conclusion$ of the rule with the maximum correlation\label{algo:inference_variant_pbre_6}
  }
  
  }
  }
  {return} $conclusion$ for $O$ \label{algo:inference_variant_pbre_7}\;
  
 \caption{extracted rules reasoner}
\label{algo:extracted_rules_reasoner}
\end{algorithm*}

\section{Decision maker}
\label{sec:decision_maker}

The "decision maker" receives the propositions from three mechanisms in the "state proposition" module. Then, it selects the final states for the corresponding actuators. Meanwhile, it sends the message to the "rule management" module specifying the propositions of which mechanisms are selected, together with the rewards sent from the "machine learning-based system", to decide whether to extract rules or delete the extracted rule.


To select the final proposition, the decision maker favors the propositions made by the existing rules since these rules are created by the inhabitant, thus ensuring the inhabitant's satisfaction. Suppose the existing rules cannot make propositions, or the propositions have conflicts. Then, the decision maker selects the propositions made by the extracted rules because the rules are extracted from the machine learning-based system only when rewards of all services are positive, ensuring the higher accuracy of the extracted rules than the machine learning-based system. Finally, if no extracted rules can make propositions, the decision maker selects the propositions made by the machine learning-based system.

\section{Working process of HKD-SHO}
\label{sec:working_process_of_hkd-sho}

\begin{figure}[t!]
  \centering
  \includegraphics[width=\linewidth]{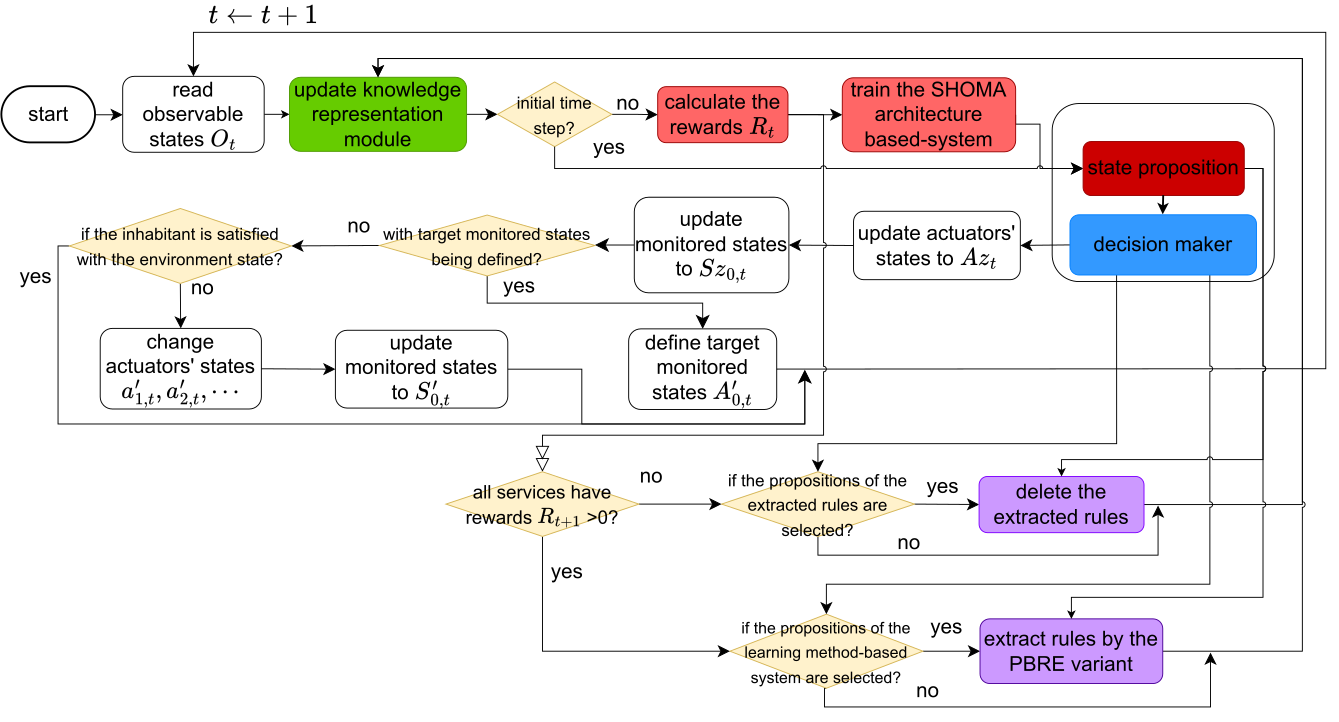}
  \caption{Detailed working process of HKD-SHO}
    \label{fig:working_process_of_HKD_SHO}
\end{figure}

The working process of HKD-SHO is in Fig.\ref{fig:working_process_of_HKD_SHO}: When the system starts, the observable states $O_t$, including sensor and actuator states, are used to update the knowledge representation module.
Then, the system asks if the current time step is the initial one. If yes, the updated knowledge is sent to the "state proposition" module, and the "decision maker" module selects final states for the actuators. Otherwise, SHOMA computes the rewards $R_{t}$ where $R_{t}$ relate to the activities of the previous iteration rather than the rewards $R_{t+1}$ for the current iteration. 

Thus, when $R_{t}$ are sent to the "rule management" module, they are used to potentially activate the rule extraction or deletion considering the activities of the previous time step $t-1$. The two white triangle arrows mean that the diamond condition is waiting for the rewards calculated in the next time step to match the activities of the current time step. The transition of the previous time step for each service $z$ is $d^z_{t-1}=\{O^z_{t-1},{Az}^z_{t-1},{s\mathsmaller{z}}^z_{0,t-1},O^z_{t},r^z_{t}\}$.

Then, the SHOMA system is trained on the updated transition dataset. The updated SHOMA system, the updated extracted rules, and the existing rules propose actuators' states within the "state proposition" module, and the "decision maker" selects the final states for the actuators. In addition to proposing the final actuators' states ${Az}_t$, the "decision maker" also sends information to the "rule management" module. The information specifies whether ${Az}_t$ is proposed by the extracted rules reasoner or the SHOMA system and includes the values of ${Az}_t$. Moreover, the "state proposition" module transmits the input states of the SHOMA system to the "rule management" module. 

The information from the "decision maker" is used together with the rewards $R_{t+1}$ computed in the next time step $t+1$ to decide whether to extract rules for the current time step $t$, delete the extracted rules used to propose actuators' states at the current time step $t$, or do nothing. The actuators change their states to the final selected states ${Az}_{t}$. The changes in the actuators' states cause the monitored states to update to ${S{z}}_{0,t}$. Subsequently, HKD-SHO asks whether the inhabitant is satisfied with ${S{z}}_{0,t}$ to calculate $R_{t+1}$ in the next time step. 

If the inhabitant is dissatisfied with the updated monitored states, he can set target values $A_{0,t}^\prime$ for the monitored states. Alternatively, he can adjust the actuators' states to $A_t^\prime$. However, the second method is primarily applicable in real-world deployment of HKD-SHO, as it is challenging to calculate inverse equations in simulations to obtain $A_{0,t}^\prime$ based on input environment states for achieving preferred monitored states values. This complexity intensifies with intricate preferences or a large number of associated actuators. If the inhabitant is content, no action is taken.


The system comes to the next time step $t+1$, and repeats the above process. During the next time step $t+1$, the rewards $R_{t+1}$ are calculated and sent to the "rule management" module. Suppose that $R_{t+1}$ are not positive for all services, and the message from the "decision maker" states that the propositions of the extracted rules consist of ${A}_{t+1}$. In this case, the extracted rules used to make the propositions at time step $t$ are deleted from the "knowledge representation" module. However, suppose $R_{t+1}$ are positive for all services, and the propositions of the SHOMA system are parts of ${A}_{t+1}$. In this case, a rule is extracted from the SHOMA system using the PBRE variant, and the extracted rule is stored in the "knowledge representation" module.

\section{Comparative experiment}
\label{sec:simulated_experiment}

To evaluate HKD-SHO, we compared it with four other systems. The first is a random system, where each service randomly proposes actuators' states. The second is the selected SHOMA. The third is the HKD-SHO without extracted rules - without the extracted rules reasoner, the "rule management" module, and the extracted rules database, while the remaining components operate similarly to HKD-SHO. The fourth is the HKD-SHO without existing rules - without the existing rules reasoner, and the existing rules database, while the remaining components operate similarly to HKD-SHO. The comparison is conducted in a simulated environment with two or three services: a light intensity service, a temperature service, and an air quality service. The subsequent section details the simulation of these services by describing the evolution of relevant variables.

\subsection{Simulated environment}

The involved variables of the simulated environment are: \begin{enumerate*}[label=(\arabic*),noitemsep] 
\item ${us}$: inhabitant state.
\item ${le/te/ae}$: outdoor light intensity/temperature/air quality.
\item ${lr/tr/ar}$: indoor light intensity/temperature/air quality.
\item ${lp}$: lamp state.
\item ${cur}$: curtain state.
\item ${ac}$: air conditioner state.
\item ${win}$: window state.
\item ${ap}$: air purifier state.
\item ${wct}$: working duration for window and curtain.
\item ${act}$: air conditioner working duration.
\item ${apt}$: air purifier working duration.
\end{enumerate*}
${lr},{tr}$ and ${ar}$ are the monitored states that the three services respectively attempt to adjust.

\subsubsection{Light intensity service}

The light intensity service takes $us$ and $le$ as input and selects $lp$ and $cur$ as output. The selected $lp$ and $cur$ are used to change $lr$. Then, a reward $r_{light}$ is calculated within the RL algorithm of the SHOMA architecture to describe whether the updated $lr$ corresponds to the inhabitant's preference. Next, the RL algorithm updates its parameters based on the historical transitions through the training process. Each transition includes the current environment states ${us}_t$ and ${le}_t$, the proposed ${lp}_t$ and ${cur}_t$, the updated ${lr}_{t+1}$, and the reward $r_{{light}_{t+1}}$. The updated RL algorithm repeats the above process afterward.  

 In our study, ${us}_t$ is randomly generated at time step $t$ by following the uniform distribution: ${us}_t={U}_{int}(0,n_{us})$, where $n_{us}$ is the total number of possible states of the inhabitant. $le$ in one day respects a Gaussian distribution \cite{ilyas2012impact,juraimi2009influence}: ${le}_t=\mathcal{N}(amplititude=600,avg=12,stddev=3)+5\cdot U(0,1)$ with some noise being modeled by a uniform distribution. ${lr}_t$ can be calculated in:
 \begin{equation}
{
    \centering
    \label{eq:indoor_light_intensity}
    \begin{aligned}
    {lr}_{t+1}=\beta\times {lp}_{t}+{le}_{t}\times {cur}_{t}
    \end{aligned}
    }
\end{equation}
where $lp\in \{0,1,2,3,4\}$ being the lamp possible value, $\beta=100$ is the light intensity that each level of the lamp can provide, and ${cur}_t\in \{0,1/2,1\}$.

\subsubsection{Temperature service}

The temperature service adjusts its monitored state $tr$ by controlling $ac, win, cur, {act}$ and $wct$. Its working process is similar to that of the light intensity service, while each transition contains the current ${us}_t, {te}_t, {tr}_t$, the proposed ${ac}_t, {win}_t, {cur}_t, {act}_t, {wct}_t$, the updated ${tr}_{t+1}$ and $r_{{temp}_{t+1}}$.

${te}_t$ can be simulated in a trigonometric function \cite{benth2012critical} by: ${te}_t=A\cdot cos(B\cdot(x-D))+C$,
where $A=-7, B=\pi/12, C=19$ and $D=4$ and $x$ is the time with unit $hour$ at time step $t$. The relation between $x$ and $t$ is: $x=t\cdot5/60$ as the time interval between $t$ and $t+1$ is 5 minutes. The possible values of each non mentioned actuator are:
\begin{enumerate*}[label=(\arabic*)] 
\item ${ac}_t\in \{0,-1,1\}$ with $0$ for off, $1$ for heating, and $-1$ for cooling;
\item ${act}_t, {wct}_t \in \{i/10 \textit{ for }i \in \{0,50\}\}$.
\end{enumerate*}
${tr}_{t+1}$ is updated in:
\begin{equation}
{
    \centering
    \label{eq:temp_final_temperature}
    \begin{aligned}
    {tr}_{t+1}=\frac{Q_{ac,t}+Q^{heat}_{loss,t}}{C_p\cdot \rho\cdot V}\pm {tr}_{t}, 
    \end{aligned}
    }
\end{equation}
where "+" and "-" are respectively heating and cooling, $Q_{ac,t}$ is the energy produced by the air conditioner after having worked ${act}_t$ hour, $Q^{heat}_{loss,t}$ is the heat exchanged between the indoor and outdoor through the window and the curtain, $C_p$ is the specific heat of the air, and $\rho$ is the air density on average. The detailed process of acquiring Eq.\ref{eq:temp_final_temperature} can be found in \cite{qiu2022reinforcement}.

\subsubsection{Air quality service}

The air quality service adjusts $ar$ by controlling $ap$, $win$, $cur$, $apt$ and $wct$. Its working process is the same as those of the above services, while each transition contains the current environment states ${us}_t, {ae}_t, {ar}_t$, the proposed ${ap}_t, {win}_t, {cur}_t, {apt}_t, {wct}_t$, the updated ${ar}_{t+1}$ and $r_{{air}_{t+1}}$. 

${ae}_t$ is simulated by modeling the atmospheric carbon dioxide dataset\footnote{The dataset can be downloaded from: \url{https://we.tl/t-Lbj5PxK2bF}} from quasi-continuous measurements on American Samoa \cite{Thoning2021}; ${ap}_t\in \{0,60,170,280,\allowbreak 390,500\}$ where $0$ means turning off, and other values represent the cubic meter air flow rate ($m^3/h \text{ or } CMH$) of the air purifier; ${apt}_t \in \{i/10 \textit{ for }i \in \{0,50\}\}$;
according to \cite{ansanay2013estimating}, ${ar}_{t+1}$ can be obtained through:
\begin{equation}
{ \small
    \centering
    \label{eq:final_indoor_air_quality}
    \begin{aligned}
    {ar}_{t+1}=&{ar}_t\cdot \left(1-\frac{{ap}_t\cdot {apt}_t}{V}-\frac{L_t\cdot {wct}_t}{V}-\frac{n_{us,t}\cdot b_{us,t} \cdot \Delta x_{us}}{V}\right)+{ae}_t\cdot\frac{L_t\cdot {wct}_t}{V}\\
    &+s_{exha,t}\cdot\frac{n_{us,t}\cdot b_{us,t}\cdot \Delta x_{us}}{V}
    \end{aligned}
    }
\end{equation}
where $V$ is the room volume, $L_t$ is the air flow rate for the indoor and outdoor air circulation, $s_{exha,t}=38000(ppm)$ represents the ${CO}_2$ content in the exhaled air, $n_{us,t}$ is the number of inhabitants in the room and has value 1 in our study, and $b_{us,t}\in \{0,11.004(mg/s), 31.44(mg/s), 7.6635 (mg/s)\}$ is the ${ CO }_2$ breathing rate of the inhabitant \cite{persily2017carbon} by supposing the inhabitant is between 21 and 30 years old, and $\Delta x_{us}$ is the inhabitant's breathing time, which is a constant value and is 5 minutes between two time steps.

\subsection{Experiment metrics}
We use the following metrics to evaluate the performance of the five systems, where an accuracy indicates the number of samples for which the service(s) can correctly propose actuators' states as a percentage of the total samples: 
\begin{enumerate*}[label=(\arabic*)]
\item the accuracy of each individual service. 
\item the accuracy of all services. In other words, the accuracy that all services simultaneously and correctly propose actuators' states. It describes the inhabitant's satisfaction with the system. 
\item the average of all accuracy, which describes the system's general performance.
\end{enumerate*}

\subsection{Experiment results}


Several comparisons\footnote{The codes are available at: \footnotesize{\url{https://github.com/mingming81/HKD-SHO}} } are conducted, each involving two or three services, considering SHOMA as either RSAbA or EPbA, with or without the constraint of minimizing the use of electrical actuators like the lamp, the air conditioner, and the air purifier.

\subsubsection{Result with two services}

\begin{figure}[t!]
\centering
   \begin{minipage}[t!]{0.83\linewidth}
     \centering
     \includegraphics[width=\textwidth]{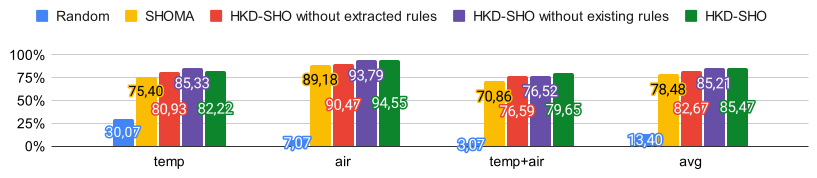}
     \caption{two services, with RSAbA and without constraint}
    \label{fig:rsaba_without_energy_saving_two_services}
    \hspace{1em}
   \end{minipage}
   \begin{minipage}[t!]{0.83\linewidth}
     \centering
     \includegraphics[width=\textwidth]{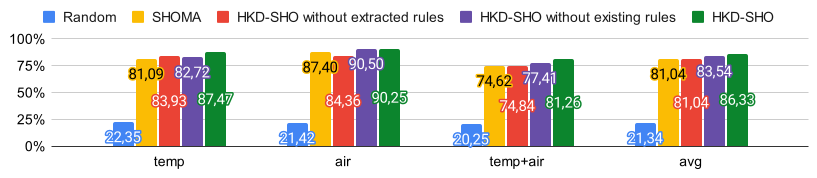}
     \caption{two services, with EPbA and without constraint}
    \label{fig:epba_without_energy_saving_two_services}
   \end{minipage}
\end{figure}

\begin{figure}[t!]
\centering
   \begin{minipage}[t!]{0.83\linewidth}
     \centering
     \includegraphics[width=\textwidth]{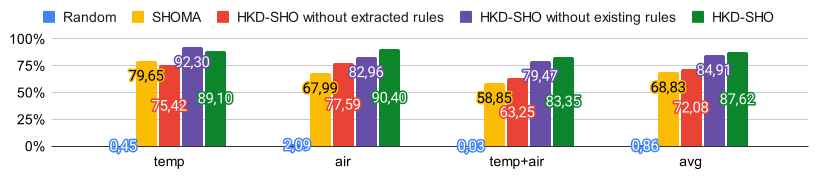}
     \caption{two services, with RSAbA and with constraint}
    \label{fig:rsaba_with_energy_saving_two_services}
    \hspace{1em}
   \end{minipage}
   \begin{minipage}[t!]{0.83\linewidth}
     \centering
     \includegraphics[width=\textwidth]{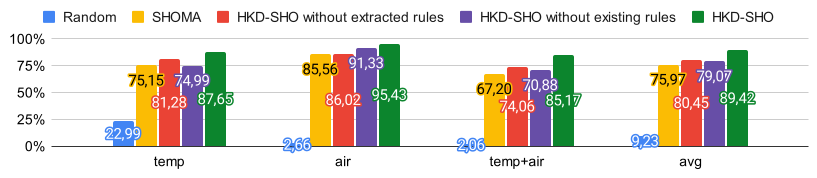}
     \caption{two services, with EPbA and with constraint}
    \label{fig:epba_with_energy_saving_two_services}
   \end{minipage}
\end{figure}

Fig.\ref{fig:rsaba_without_energy_saving_two_services} and Fig.\ref{fig:epba_without_energy_saving_two_services} show the results with two services, SHOMA being respectively RSAbA and EPbA, and no constraint being in each service. By observing the accuracy of all services "temp+air" and the average accuracy "avg", HKD-SHO based on EPbA performs better than HKD-SHO based on RSAbA. Regarding the accuracy of individual services, we cannot tell which is better between HKD-SHO based on RSAbA and EPbA. Also, by observing all metrics in each experiment, we generally have the performance order: HKD-SHO>HKD-SHO without existing rules>HKD-SHO without extracted rules>SHOMA>Random system. Fig.\ref{fig:rsaba_with_energy_saving_two_services} and Fig.\ref{fig:epba_with_energy_saving_two_services} show the results with the constraint, and we can obtain the same results as those in Fig.\ref{fig:rsaba_without_energy_saving_two_services} and Fig.\ref{fig:epba_without_energy_saving_two_services}.

\subsubsection{Results with three services}

\begin{figure}[t!]
\centering
   \begin{minipage}[t!]{0.83\linewidth}
     \centering
     \includegraphics[width=\textwidth]{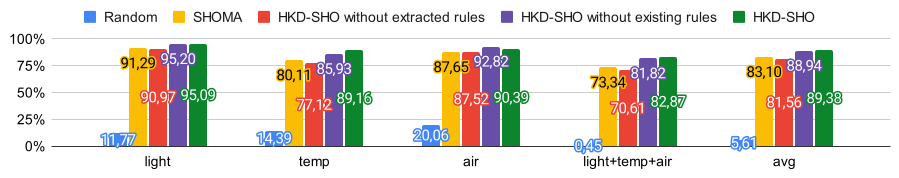}
     \caption{Three services, with RSAbA and without constraint}
    \label{fig:rsaba_without_energy_saving_three_services}
    \hspace{1em}
   \end{minipage}
   \begin{minipage}[t!]{0.83\linewidth}
     \centering
     \includegraphics[width=\textwidth]{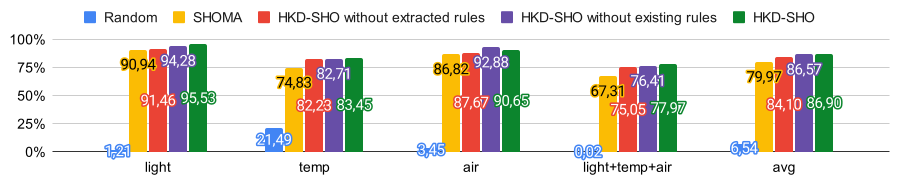}
     \caption{Three services, with EPbA and without constraint}
    \label{fig:epba_without_energy_saving_three_services}
   \end{minipage}
\end{figure}

\begin{figure}[t!]
\centering
   \begin{minipage}[t!]{0.83\linewidth}
     \centering
     \includegraphics[width=\textwidth]{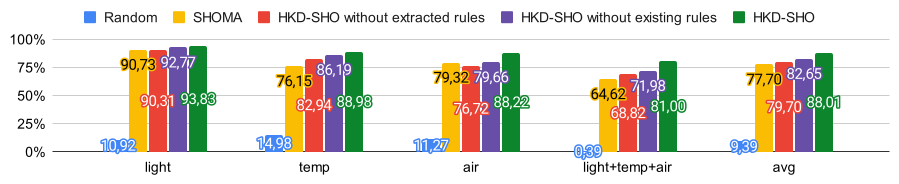}
     \caption{three services, with RSAbA and with constraint}
    \label{fig:rsaba_with_energy_saving_three_services}
    \hspace{1em}
   \end{minipage}
   \begin{minipage}[t!]{0.83\linewidth}
     \centering
     \includegraphics[width=\textwidth]{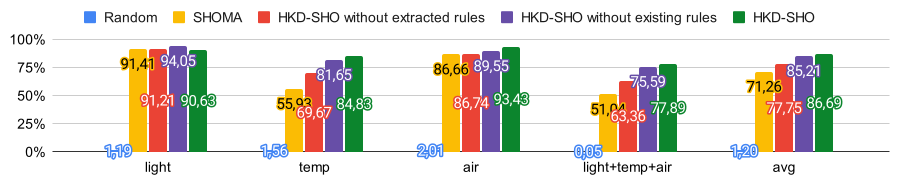}
     \caption{three services, with EPbA and with constraint}
    \label{fig:epba_with_energy_saving_three_services}
   \end{minipage}
\end{figure}

In Fig.\ref{fig:rsaba_without_energy_saving_three_services} and Fig.\ref{fig:epba_without_energy_saving_three_services}, results are presented for three services, with SHOMA being RSAbA or EPbA and without constraints. Notably, in metrics "avg" and "light+temp+air," HKD-SHO based on RSAbA outperforms HKD-SHO based on EPbA. However, the comparison for accuracy in individual services between EPbA and RSAbA-based HKD-SHO is inconclusive. The overall performance order across all metrics in each experiment generally aligns with the conclusions from the previous experiment. Results with constraints, as depicted in Fig.\ref{fig:rsaba_with_energy_saving_three_services} and Fig.\ref{fig:epba_with_energy_saving_three_services}, yield similar conclusions.

\subsubsection{Conclusion and analysis}
Through the above experiments, we can conclude that, first, regarding the metrics "the accuracy of all services" and "the average of all accuracy", when there are two services, HKD-SHO based on EPbA has better performance than HKD-SHO based on RSAbA; the result is reversed when there are three services. Second, by observing the accuracy of individual services, we cannot tell which one is better between HKD-SHO based on EPbA and RSAbA. Regarding all metrics, for each experiment, we can obtain the third to sixth conclusions: Third, the random system has the worst performance, which indirectly proves the SHOMA-based systems (systems except the random system) have learned something from the simulated experiment to adapt to the inhabitant's favorite monitored state values. Fourth, in most situations, HKD-SHO without extracted rules performs better than SHOMA, which proves that introducing existing rules contributes to improving the performance of SHOMA. Fifth, in most situations, HKD-SHO without existing rules performs better than HKD-SHO without extracted rules. This may be because there are more extracted rules than existing ones, thus contributing more to the decision-making despite the machine learning-based system. Sixth, in general situations, HKD-SHO has the best performance, proving its satisfying performance in the simulated experiment. This is because combining existing and extracted rules increases the number of rules, thus contributing more to SHOMA based on each type of rule.


\section{Conclusion}
\label{sec:conclusion}

To address the dynamism of knowledge-based approaches and the explicability of data-driven approaches during service creation, we have proposed a hybrid system HKD-SHO that can create dynamic smart home services due to the active learning of the MARL algorithm while guaranteeing that these services are explainable through rule extraction from MARL.
In perspective, we can enhance the performance of HKD-SHO by improving the sample efficiency of the machine learning-based system. After that, it is necessary to evaluate the hybrid system by deploying it in a real smart home.


%
%
%
\bibliographystyle{splncs04}
\bibliography{mybibliography}
%




\end{document}